# 스테레오 비전을 이용한 실시간 도로 표면 재구성 방법 연구

기미레 디팍, 김병준, 김동훈, 정성환
IT응용연구센터, 한국전자기술연구원

# Study on Real-Time Road Surface Reconstruction Using Stereo Vision

Deepak Ghimire, Byoungjun Kim, Donghoon Kim, and SungHwan Jeong
IT Application Research Center, Korea Electronics Technology Institute

**Abstract** – Road surface reconstruction plays a crucial role in autonomous driving, providing essential information for safe and smooth navigation. This paper enhances the RoadBEV [1] framework for real-time inference on edge devices by optimizing both efficiency and accuracy. To achieve this, we proposed to apply Isomorphic Global Structured Pruning to the stereo feature extraction backbone, reducing netwok complexity while maintaining performance. Additionally, the head network is redesigned with an optimized hourglass structure, dynamic attention heads, reduced feature channels, mixed precision inference, and efficient probability volume computation. Our approach improves inference speed while achieving lower reconstruction error, making it well-suited for real-time road surface reconstruction in autonomous driving.

## 1. Introduction

Autonomous driving systems rely heavily on accurate and real-time road surface perception for safe and efficient navigation. Understanding road geometry, particularly elevation variations such as bumps and potholes, is crucial for vehicle motion planning, ride comfort, and safety [2]. While LiDAR-based approaches provide high-precision road surface reconstruction, their high cost and power consumption make them impractical for many applications, especially in cost-sensitive scenarios [3]. Vision-based methods, leveraging monocular and stereo cameras, offer a more economical alternative by estimating depth maps and reconstructing road geometry. However, traditional depth estimation methods struggle with perspective distortions, making fine-grained road surface reconstruction challenging.

Recent advancements in Bird's Eye View (BEV) representations have demonstrated significant improvements in 3D perception tasks, including object detection and segmentation. Among these, RoadBEV [1] has emerged as a promising framework for road surface reconstruction, surpassing conventional stereo depth estimation methods while maintaining a relatively simple architectural design. By transforming image features into BEV space, RoadBEV effectively captures local elevation variations, leading to superior reconstruction accuracy. However, despite its advantages, RoadBEV remains computationally expensive, making it unsuitable for real-time deployment on edge devices.

To address this limitation, we propose a restructured and optimized version of RoadBEV, maintaining its performance while significantly improving efficiency. The primary challenge lies in the computational complexity of both the backbone and the head network. The existing structure, while effective, involves high memory and processing demands, limiting its applicability in resource-constrained environments. One potential solution is model compression, specifically structured pruning, which allows for the reduction of redundant computations while preserving network expressiveness. Structured pruning has been widely studied in deep learning model optimization, demonstrating effectiveness in reducing model size and improving inference speed. Several works, such as DepGraph [4], Isomorphic Pruning for Vision Models [5], and One-cycle Structured Pruning with Stability Driven Structure Search [6], have explored various techniques to prune model structures while maintaining or even enhancing performance. Inspired by these approaches, we apply Isomorphic Global Structured Pruning to the RoadBEV backbone, systematically removing redundant parameters while retaining its critical spatial representation capabilities.

Furthermore, the head network in RoadBEV is nearly as complex as the backbone, making it another computational bottleneck. To enhance efficiency, we redesign the head architecture by incorporating an optimized hourglass structure, dynamic attention heads, reduced feature channels, mixed precision inference, and efficient probability volume computation. These modifications streamline processing while preserving essential features required for accurate road surface reconstruction.

The proposed enhancements significantly reduce computational overhead, making RoadBEV suitable for real-time deployment on edge devices without sacrificing performance. Our optimized model achieves faster inference while maintaining or even improving reconstruction accuracy, thereby bridging the gap between high-performance road surface perception and practical deployment in autonomous driving systems.

## 2. Proposed Real-Time RoadBEV Framework

The original RoadBEV [1] provides a robust foundation for road surface reconstruction. However, to enable real-time inference on edge devices with limited computational resources, we propose a series of optimizations. Our framework, Real-Time (RT)-RoadBEV, incorporates advancements in feature extraction, network design, and inference efficiency, leading to faster processing times without compromising accuracy. Figure 1 shows the overall structure, with (a) depicting the original RoadBEV [1] and (b) showing our real-time RT-RoadBEV.

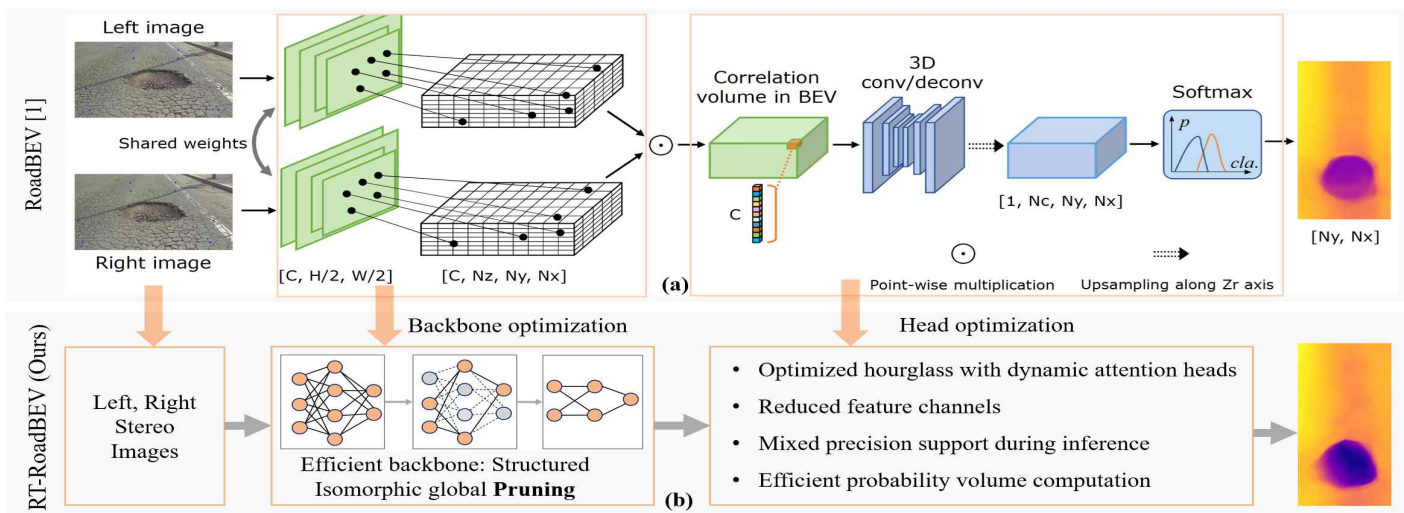

⟨Figure 1⟩ Overview of the proposed framework. (a) Original RoadBEV [1], and (b) Our enhanced RT-RoadBEV.

### 2.1. Feature Extraction Backbone Optimization

The original RoadBEV framework uses a manually redesigned EfficientNet-B6 [12] model for feature extraction. Since both the left and right stereo images pass through this backbone, optimizing it for real-time inference is critical. Instead of relying on manual design, we propose the use of structured pruning to optimize the backbone network. Structured pruning methods, such as those in [4-6], can generate task-specific, optimal structures from base models. Among these, we choose Isomorphic Global Pruning [5] as it is a state-of-the-art method well-suited for our task. Unlike model quantization, which is hardware-dependent, structured pruning is a flexible, hardware-independent approach. Specifically, Isomorphic Global Pruning ranks and prunes network sub-structures based on a pre-defined importance criterion, ensuring pruning preserves model accuracy. This technique effectively reduces computational complexity by eliminating less important components, without compromising backbone performance.

### 2.1 Head Network Redesign

Our complexity analysis of the original RoadBEV framework revealed that the head network is as computationally intensive as the backbone. Therefore, we propose optimizing the head network alongside the backbone. While end-to-end pruning could be applied to the entire framework, maintaining accuracy during optimization is crucial. As a result, we implemented several strategies to optimize the head network independently of the backbone. These include redesigning the head network with an optimized hourglass structure, incorporating dynamic attention heads, reducing feature channels, using mixed precision inference, and improving probability volume computation. With these enhancements, we achieved nearly a 50% reduction in the head network's computation time while maintaining accuracy.

## 3. Experiments
### 3.1 Dataset and Implementation Details

We conduct our experiments on the Road Surface Reconstruction Dataset (RSRD) [11], following the exact preprocessing and training pipeline of RoadBEV [1] to ensure a fair comparison. As such, we do not provide detailed descriptions of these settings but refer readers to RoadBEV [1] and RSR [11] for more information. The dataset consists of 2,800 stereo image pairs, each with corresponding LiDAR point clouds and calibration files. From this, we sample 1,210 training pairs and 371 test pairs. The LiDAR point cloud data serves as the ground truth for depth estimation during training. Our implementation is built on PyTorch, and all experiments are conducted on an NVIDIA RTX 4080 GPU. We use a batch size of 2 and an initial learning rate of 5e-4, with optimization strategies adapted for efficient training and inference.

<Table 1> Performance comparison of RT-RoadBEV with state-of-the-art stereo matching and RoadBEV [1] methods.

| Method | Abs. err. (cm) | RMSE (cm) | >0.5cm (%) | FPS |
|---|---|---|---|---|
| IGEV-Stereo [8] | 0.651 | 0.797 | 49.5 | 4.6 |
| CFNet [9] | 0.647 | 0.760 | 50.8 | 6.8 |
| DVANet [11] | 0.546 | 0.685 | 40.9 | 8.7 |
| RoadBEV [1] | 0.503 | 0.609 | 37.0 | 7.9 |
| RT-RoadBEV | 0.497 | 0.603 | 36.93 | 10.9 |

### 3.2 Results

Table 1 presents a comparative analysis of RT-RoadBEV, RoadBEV, and other stereo depth estimation methods. Our approach achieves higher reconstruction accuracy while significantly reducing computational cost. A key optimization in RT-RoadBEV is the head module, which cuts inference time from 70ms to 35ms, reducing computational overhead by 50%. While backbone pruning is not yet fully implemented, the current optimizations already provide substantial efficiency gains. RT-RoadBEV surpasses all baselines in absolute error, RMSE, and >0.5cm error rate, maintaining competitive accuracy despite reduced complexity. With optimized FLOPs and parameter count, we achieve about 11 FPS, with further improvements possible through backbone refinement.

Figure 2 visually shows the reconstruction results obtained with our RT-RoadBEV for the test cases of road bumps and potholes. The second row illustrates the estimated elevation maps in BEV, while the third row presents the reconstructed 3D road meshes. The results demonstrate that our method maintains high reconstruction fidelity, even with reduced computation. These findings highlight the effectiveness of our architectural optimizations, making RT-RoadBEV a strong candidate for real-time road surface reconstruction in autonomous systems.

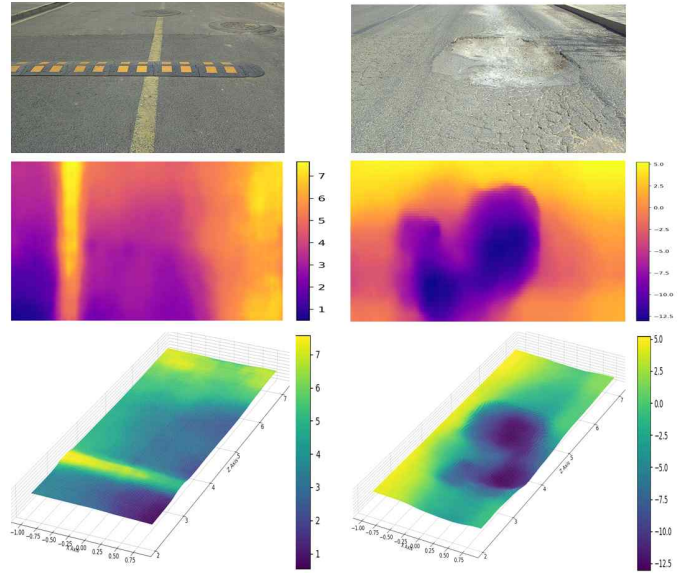

<Figure 2> Visualization of road surface reconstruction by RT-RoadBEV. From top to bottom: RGB images, estimated elevation maps, and corresponding 3D road meshes.

## 4. Summary

This paper enhances RoadBEV for real-time road surface reconstruction by improving efficiency and accuracy. We apply Isomorphic Global Structured Pruning to reduce FLOPs and parameters while maintaining performance. The head network is redesigned with an optimized hourglass structure, dynamic attention heads, and mixed precision inference, boosting speed and accuracy. Experiment shows RT-RoadBEV achieves superior accuracy with lower computational cost. The optimized head module reduces inference time from 70ms to 35ms, reaching overall 11 FPS, making it suitable for real-time autonomous driving.

### Acknowledgment

본 논문은 2025년도 산업통상자원부의 재원으로 한국산업기술기획평가원(KEIT) 자동차산업기술개발사업 "상용 특수 및 작업보조 차량의 자율 주행 협업 제어 플랫폼 개발(20018906)" 사업의 지원을 받아 수행된 연구 결과임.